\documentclass[10pt,twocolumn,letterpaper]{article}

\usepackage{iccv}
\usepackage{times}
\usepackage{epsfig}
\usepackage{graphicx}
\usepackage{amsmath}
\usepackage{amssymb}

\usepackage{enumitem} 
\usepackage{verbatim} 
\usepackage{xcolor} 
\usepackage{subfigure} 
\usepackage{multirow}
\usepackage{multicol}
\usepackage{bigstrut}
\usepackage{booktabs}
\usepackage{comment}

\newcommand{\ieno}{\textit{i}.\textit{e}.} 
\newcommand{\egno}{\textit{e}.\textit{g}.} 
\newcommand{\etcno}{\textit{etc}} 

\usepackage{array}
\newcolumntype{C}[1]{>{\centering\arraybackslash}m{#1}}
\newcolumntype{R}[1]{>{\raggedleft\arraybackslash}m{#1}}
\newcolumntype{P}[1]{>{\raggedright\arraybackslash}p{#1}}
\newcolumntype{M}[1]{>{\centering\arraybackslash}m{#1}}

\usepackage[pagebackref=true,breaklinks=true,letterpaper=true,colorlinks,bookmarks=false]{hyperref}

\iccvfinalcopy 

\def\iccvPaperID{2577} 

\ificcvfinal\fi

\begin{document}

\title{Disentanglement-based Cross-Domain Feature Augmentation for Effective \\ Unsupervised Domain Adaptive Person Re-identification} 

\author{Zhizheng Zhang{$^1$}\thanks{This work was done when Zhizheng and Kecheng were interns at Microsoft Research Asia.} \quad 
Cuiling Lan$^2$ 
\quad 
Wenjun Zeng$^2$ 
\quad 
Quanzeng You$^2$
\quad
Zicheng Liu$^2$
\\
Kecheng Zheng$^1$
\quad
Zhibo Chen$^{1}$ 
\and 
\normalsize
	$^1$University of Science and Technology of China \qquad $^2$Microsoft Research\\
	{\tt\small \{zhizheng, zkcys001\}@mail.ustc.edu.cn} \quad {\tt\small chenzhibo@ustc.edu.cn} \\ {\tt\small \{culan, wezeng, quanzeng.you, zliu\}@microsoft.com}
}

\maketitle
\ificcvfinal\thispagestyle{empty}\fi

\begin{abstract}
    Unsupervised domain adaptive (UDA) person re-identification (ReID) aims to transfer the knowledge from the labeled source domain to the unlabeled target domain for person matching. One challenge is how to generate target domain samples with reliable labels for training. To address this problem, we propose a Disentanglement-based Cross-Domain Feature Augmentation (DCDFA) strategy, where the augmented features characterize well the target and source domain data distributions while inheriting reliable identity labels. Particularly, we disentangle each sample feature into a robust domain-invariant/shared feature and a domain-specific feature, and perform cross-domain feature recomposition to enhance the diversity of samples used in the training, with the constraints of cross-domain ReID loss and domain classification loss. Each recomposed feature, obtained based on the domain-invariant feature (which enables a reliable inheritance of identity) and an enhancement from a domain specific feature (which enables the approximation of real distributions), is thus an ``ideal" augmentation. Extensive experimental results demonstrate the effectiveness of our method, which achieves the state-of-the-art performance.

\end{abstract}

\section{Introduction}

Person re-identification (ReID) aims to identify the same person across different locations, time instances, and cameras. This technique is potentially useful for many applications such as tracking people for smart retail and finding missing children. In real-world applications, when a trained model is deployed to a new environment, it may suffer from severe performance drop due to the domain gap between the data from the new environment and the model's training data. Collecting and manually annotating data from new environments to fine-tune the model can alleviate this problem but is costly. A much cheaper and more attractive solution is to employ an unsupervised domain adaptive technique, which attempts to exploit the knowledge from labeled source domain and unlabeled target domain to achieve better performance in the new environment.

\begin{figure}[t]
	\begin{center}
		\includegraphics[width=0.96\linewidth]{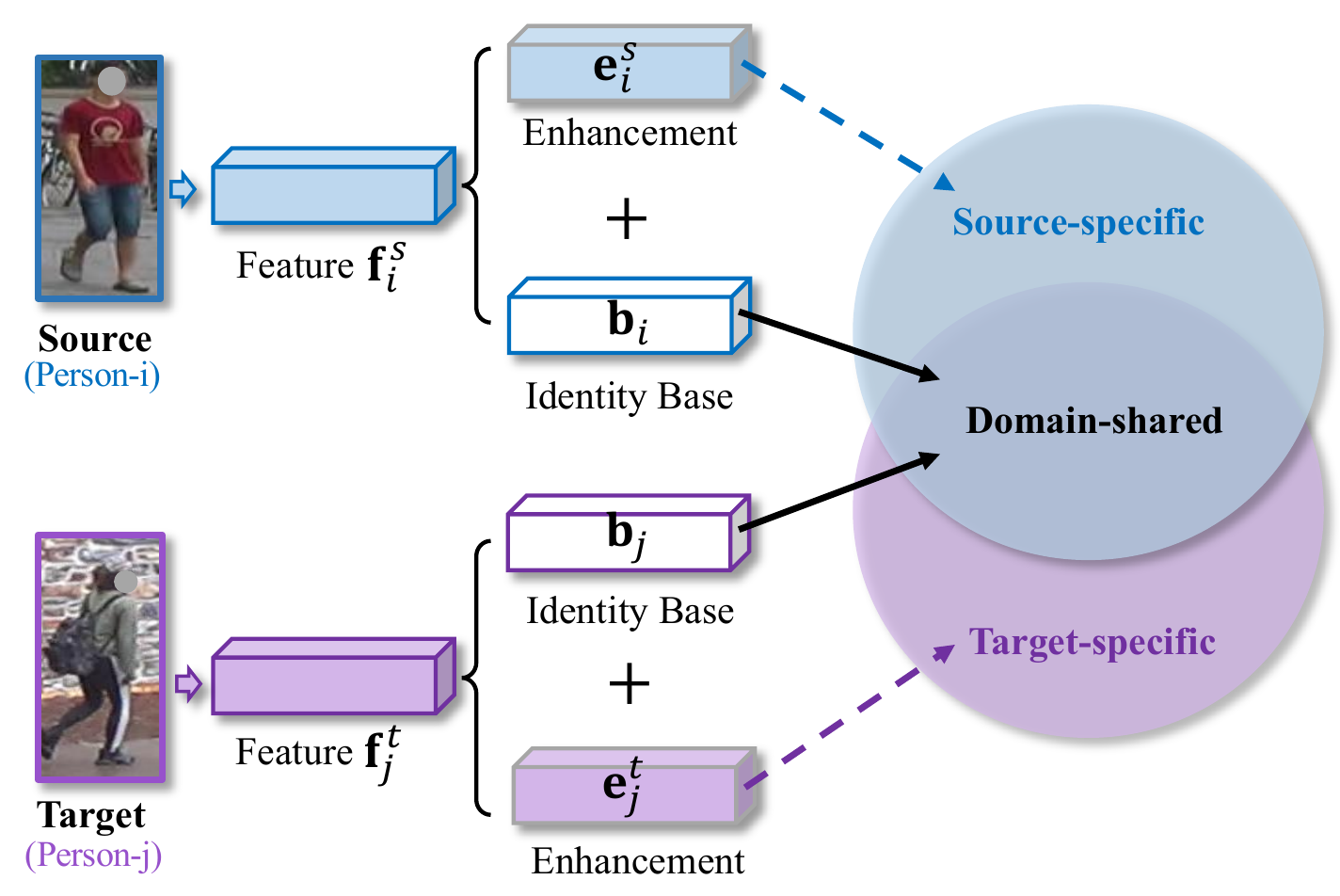}
	\end{center}
	\vspace{-2mm}
	\caption{Illustration of domain-shared/invariant and domain-specific components. We represent the ID feature of each person in a scalable way with domain-shared feature as base and domain-specific feature as enhancement.}
	\label{fig:keyidea}
	\vspace{-2mm}
\end{figure}

Recently, many approaches are proposed for unsupervised domain adaptive person ReID. Clustering-based approaches \cite{fan2018unsupervised,zhang2019self,fu2019self,ge2020mutual,yu2019unsupervised,jin2020global,zhao2020unsupervised} are popular which assign pseudo labels for the target domain samples by applying different clustering methods. Their common pipeline is to train a model on labeled source domain and then fine-tune the model on the target domain with pseudo labels.
However, because of the existing domain gap between source and target datasets, the psuedo-labels obtained/assigned through clustering usually contain noisy/unreliable labels. Such unreliable labels would mislead the feature learning and prevent the achievement of high domain adaptation performance. \textbf{The lack of reliable labels for the target domain samples is one big challenge for UDA person ReID.} Some methods tend to alleviate the influence of noisy pseudo labels by exploiting mutual learning \cite{ge2020mutual,zhai2020multiple,zhao2020unsupervised} or/and ignoring the outliers~\cite{zhao2020unsupervised,ge2020self} for training. For example, NRMT \cite{zhao2020unsupervised} maintains two networks to perform collaborative clustering and mutual instance selection, which reduces the chance of fitting to noisy instances by using the mutual supervision and the reliable instance selection in training. \textbf{In contrast, we address the problem from a new perspective, \ieno, by generating target domain samples with reliable labels for training.} 

Unlike previous methods that generate samples based on complicated generative models, \ieno, Generative Adversarial Networks \cite{wei2018person,deng2018image,liu2019adaptive,ge2020structured}, we propose a Disentanglement-based Cross-Domain Feature Augmentation (DCDFA) strategy to generate ``ideal" augmented features, which characterize well the target and source domain data distributions while inheriting reliable identity labels. Particularly, we disentangle each extracted sample feature into a robust domain-invariant/shared feature and a domain-specific feature (see \figurename~\ref{fig:keyidea}). As illustrated in \figurename~\ref{fig:framework}, we then perform feature recomposition for a domain-invariant/shared feature in one domain and a domain-specific feature in the other domain, to enhance the diversity of sample features used in the training. We encourage the disentanglement and enable the supervision on the recomposed features with the constraints of cross-domain ReID loss and domain classification loss. Such recomposed features act as ideal augmentation, which enables reliable inheritance of identity (thanks to the use of domain-invariant features as base) and approximates the real distributions (thanks to the use of domain-specific features of real samples as enhancement). 

Note that our method is very different from previous disentanglement-based UDA methods \cite{Huang2020aaai,liu2020domain,lin2018multi,wilson2020survey,ben2007analysis,ganin2015dann}, which aim to learn domain-invariant features with reduced feature distribution discrepancy between the source and target domains. We intend to achieve ``ideal" feature augmentation for effective robust/generalizable feature learning based on disentanglement, where both domain-invariant feature and domain-specific feature are made full use of.       

We summarize our main contributions as follows:

\begin{itemize}[noitemsep,nolistsep,leftmargin=*]
\item We address the challenge of lack of reliable identity labels for the target domain samples in UDA from the perspective of generating diverse target domain samples which approximate the real data distribution well and have reliabel labels.

\item We propose an effective Disentanglement-based Cross-Domain Feature Augmentation (DCDFA) strategy, which is capable of providing diverse ``ideal" augmented features for training. Particularly, we disentangle a feature into a domain-invariant feature and a domain specific feature, then perform cross-domain recomposition to generate augmented features for training. The recomposed features not only enable reliable inheritance of identity but also approximate the real distributions.
\end{itemize}

Our proposed DCDFA brings significant improvements, and helps to achieve state-of-the-art performance on top of a strong baseline. 

\section{Related work}

\subsection{Unsupervised Domain Adaptive Person ReID}
In recent years, many deep-learning based approaches are designed for unsupervised domain adaptative person ReID. They can be grouped into three categories.

\noindent\textbf{Image-style transfer based methods} transfer source domain labeled images to match the styles of target domain images, which can be used to fine-tune the models~\cite{wei2018person,deng2018image,liu2019adaptive,ge2020structured}. The performance of these approaches is usually limited by the quality of the translated images, which is still not satisfactory. Besides, they usually leverage GAN networks, \egno, CycleGAN \cite{zhu2017unpaired} for the translation, which increases the training complexity.
In our work, we do not need to transfer the images. Instead, based on recomposition of disentangled features, we obtain augmented features of source and target domains with \emph{reliable} identity labels for training.

\noindent\textbf{Clustering-based pseudo labeling methods} are popular and have achieved impressive performance~\cite{fan2018unsupervised,zhang2019self,fu2019self,yu2019unsupervised,ge2020mutual,zhai2020multiple,jin2020global,zhao2020unsupervised}. They usually pre-train the model using labeled source samples for learning good feature representations. To fine-tune the network using target samples, they predict their pseudo labels based on clustering results using the extracted features. The performance of these methods is hindered by the noises in pseudo labels caused by the feature extractor affected by domain gaps and the clustering itself. To alleviate the influence of noisy/unreliable pseudo labels, mutual learning among several collaborative peer networks \cite{ge2020mutual,zhai2020multiple} is proposed to refine pseudo labels with each other. Collaborative clustering and mutual instance selection \cite{zhao2020unsupervised} is introduced to alleviate the effects of noise. 

Our proposed DCDFA method is orthogonal with such clustering-based approaches. Instead of focusing on refining pseudo labels as in \cite{ge2020mutual,zhao2020unsupervised}, we generate target domain sample features with reliable labels for training, based on disentangled features. Our network is simple in design, which does not need two or more networks as in mutual learning. We demonstrate the effectiveness of our DCDFA on top of clustering-based methods.

\noindent\textbf{Learning domain-invariant feature based methods} intend to learn domain-invariant features by using adversarial learning \cite{Huang2020aaai,liu2020domain} or explicitly reducing the feature distribution discrepancy between source and target domain measured by some metrics, \egno, Maximum Mean Discrepancy (MMD) \cite{lin2018multi}. This idea is widely explored/used in UDA classification \cite{lin2018multi,wilson2020survey,ben2007analysis,ganin2015dann}, which can be considered as disentanglement-based approaches in a broad sense.
However, domain-invariant features in general cannot include all the discriminative information, whereas there is still discriminative information in the domain-specific feature but under-explored. 
Unlike \cite{lin2018multi,Huang2020aaai,liu2020domain} which aim to learn domain-invariant features and use only them in inference, we leverage feature disentanglement to facilitate effective feature augmentation for learning generalizable/robust features, where both domain-invariant feature and domain-specific feature are made full use of.

\subsection{Data Augmentation}
Data augmentation aims to to increase the effective size of training data, where region-level~\cite{devries2017improved,zhong2020random} and image-level \cite{shorten2019survey,cubuk2018autoaugment} augmentation are widely used. Region-level augmentation, \egno, cutout~\cite{devries2017improved} and random erasing~\cite{zhong2020random}, modifies local rectangular regions of the input images to generate partially occluded data samples for training. Image-level augmentation exploits the invariance properties of images by applying transformation on the images, such as rotation, flipping, color jittering while preserving the labels. However, the augmentations are usually manually designed and heuristically chosen, where there is no guarantee that they are beneficial and expand the data space properly. AutoAugment~\cite{cubuk2018autoaugment} applies reinforcement learning to search optimal compositions of transformations, where the computational burden is heavy even. Its variants RandAugment~\cite{cubuk2020randaugment}, Fast AutoAugment~\cite{lim2019fast} try to alleviate this. Some methods exploit GANs and VAEs for data augmentation, with the cost of increasing the complexity of the designs \cite{shorten2019survey}.  
In this work, we propose a cross-domain feature augmentation strategy to enable the generation of ``realistic" features for effective UDA person ReID, which enables the inheritance of reliable identity labels while preserving the target and source domain data distributions.

\begin{figure*}[t]
	\begin{center}
		\includegraphics[width=0.95\linewidth]{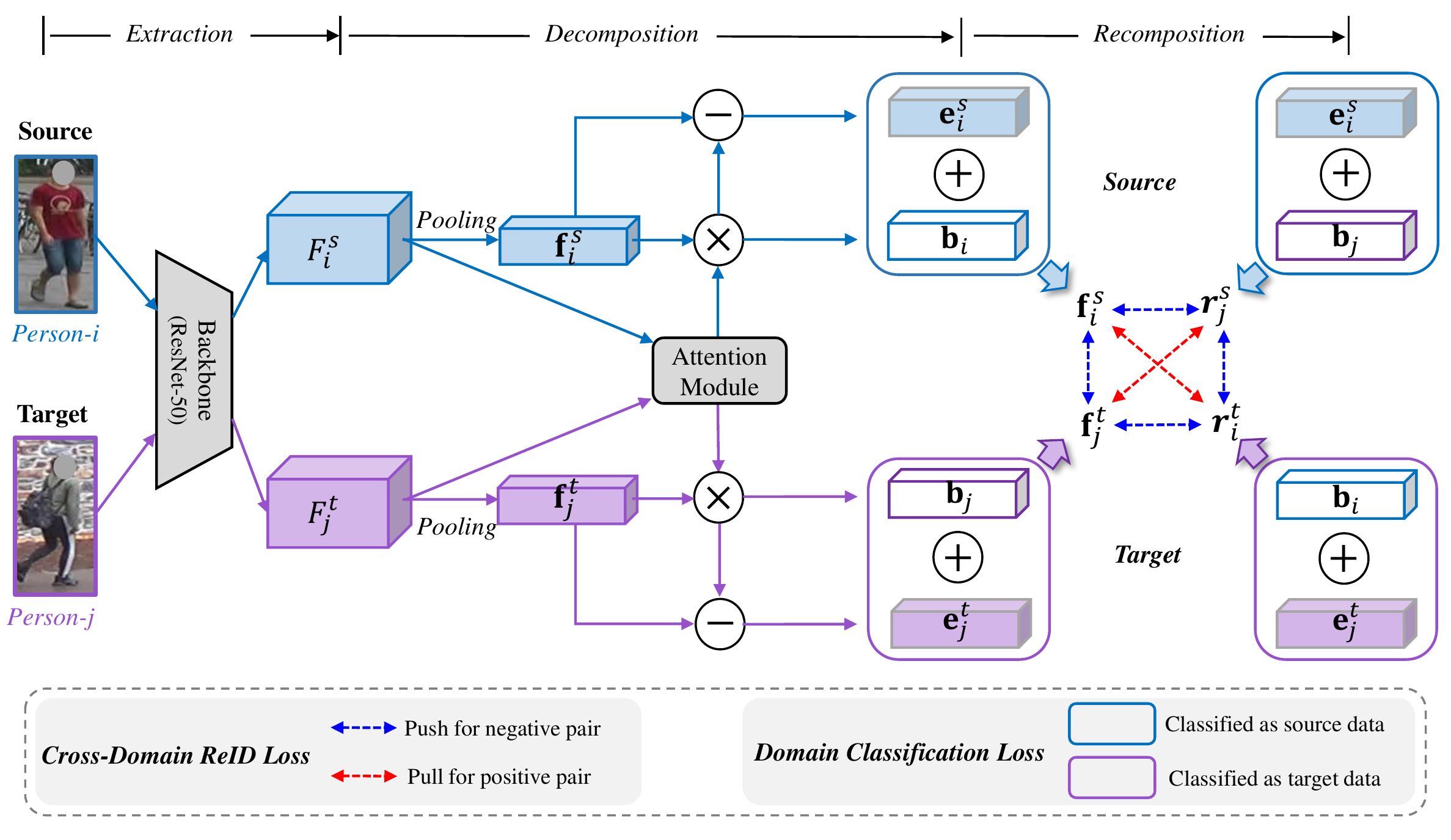}
	\end{center}
	\caption{Illustration of proposed Disentanglement-based Cross-Domain Feature Augmentation (DCDFA) for UDA person ReID, which is capable of providing diverse ``ideal" augmented features that have reliable identity labels and approximate the real data distributions well. For a source image $i$ and a target image $j$, we obtain the feature $\mathbf{f}_i^s$ and $\mathbf{f}_j^t$. Particularly, we decompose them into domain-shared base features $\mathbf{b}_i$, $\mathbf{b}_j$ and domain-specific features $\mathbf{e}_i^s$, $\mathbf{e}_j^t$. We exchange their domain-shared features to get recomposed features $\mathbf{r}_{j}^{s}$ and $\mathbf{r}_{i}^{t}$ as augmented results for training, with the constraints of cross-domain ReID loss and domain classification loss. 
	In inference, for the target image $j$, feature $\mathbf{f}_{j}^{t}$ is used for matching, where the attention module can be discarded during testing since $\mathbf{f}_{j}^{t}=\mathbf{b}_j + \mathbf{e}_j^t$.	
	}
	\label{fig:framework}
	\vspace{-2mm}
\end{figure*}

\section{Proposed method}


Unsupervised Domain Adaptive (UDA) person ReID aims to adapt the model trained on a labeled source domain $\mathcal{D}_{s}=\{(\mathbf{x}_{i}^{s}, {y}_{i}^{s})|_{i=1}^{N_{s}}\}$ to an unlabelled target domain $\mathcal{D}_{t}=\{\mathbf{x}_{i}^{t}|_{i=1}^{N_{t}}\}$, where $N_{s}$ and ${N_{t}}$ denote the numbers of samples in $\mathcal{D}_{s}$ and $\mathcal{D}_{t}$, respectively. In general, there is no overlap of identity labels between $\mathcal{D}_{s}$ and $\mathcal{D}_{t}$. 

For UDA person ReID, there are usually style differences between domains, resulting in domain gaps. Such domain gaps could be caused by the difference in environments (\egno, lighting, background, seasons, \etcno.), and capturing cameras (\egno, resolution, contrast, hue, saturation, \etcno.). For images of different domains, some information is domain transferable (\egno, person attributes, such as age, gender, clothing color, \etcno.) while some is domain-specific (\egno, clothing styles, illumination, \etcno.). 
Both of them may contain discriminative information for ReID. 

Considering the lack of reliable labels for target domain samples in training, we propose an effective Disentanglement-based Cross-Domain Feature Augmentation (DCDFA) strategy, which provides diverse and ``ideal'' augmented features for training. Particularly, as illustrated in \figurename~\ref{fig:framework}, we disentangle each sample feature into a domain-invariant feature and a domain-specific feature, then perform cross-domain recomposition to generate augmented features for training. To achieve this, the cross-domain ReID loss and domain classification loss are designed. 
The recomposed features act as ``ideal" augmentation, which not only enables reliable inheritance of identity (thanks to the use of domain-invariant features as base) but also approximates the real target distributions (thanks to the use of domain-specific features of real samples as enhancement).

In the following, we first describe the feature disentanglement in Section~\ref{subsec:fd} and the cross-domain feature recomposition for augmentation in Section \ref{subsec:fr}, respectively. We elaborate on the optimization designs which drive the network to disentangle features and perform feature augmentation in Section~\ref{subsec:optimization}. 

\subsection{Feature disentanglement}
\label{subsec:fd}
As illustrated in \figurename~\ref{fig:framework}, given a person image $\mathbf{x}$, we extract a feature map $F\!\in\!\mathbb{R}^{C \times H \times W}$ with channel number of $C$ and spatial resolution of $H \times W$, by using a Convolutional Neural Network (CNN) (\egno, ResNet-50) as the backbone. A global spatial average pooling operation is adopted to obtain a feature vector $\mathbf{f}\!\in\!\mathbb{R}^{C}$ from $F$ (\ieno, $\mathbf{f}= {\rm pool_{sa}}(F)$) as the feature representation for matching.


We aim to augment the features based on disentangled features for effective UDA. 
\emph{For UDA person ReID, an ``ideal" augmentation needs to meet two requirements: 1) the augmented target (or source) domain features should approximate the real distribution of the target (or source) domain samples; 2) the augmented features should be supplemented with reliable identity information so as to facilitate the learning process.}

One may wonder what characteristics of the disentanglement is desired to facilitate such ``ideal" augmentation.
Motivated by the two above requirements, we expect to disentangle the feature $\mathbf{f}$ into a domain-shared/invariant identity base feature $\mathbf{b}$ and a domain-specific enhancement feature $\mathbf{e}$ as:
\begin{equation}\label{eq:1}
    \mathbf{f}=\mathbf{b}+\mathbf{e},
\end{equation}
where the domain-shared feature $\mathbf{b}$, as identity base feature, predominates in identifying/recognizing the person identity; the domain-specific $\mathbf{e}$, as its name implies, constitutes the domain-specific information and acts as enhancement to the former. \textbf{The domain-shared features, together with the domain-specific features of the target (or source) domain, characterize the data distribution of the target (or source) domain, where the domain-shared features of different domains are exchangeable between domains without any destruction to each domain's distribution.}

Particularly, we implement the feature decomposition (\ie disentanglement) by simply using a channel-wise attention module $\mathbf{m}(\cdot)$ \cite{hu2018squeeze,woo2018cbam} as: 
\begin{equation}\label{eq:2}
    \mathbf{b} = \mathbf{m}(F)\odot \mathbf{f},\quad
    \mathbf{e} = (1-\mathbf{m}(F))\odot \mathbf{f},
\end{equation}
where $\odot$ denotes element-wise multiplication and $\mathbf{m}(\cdot) \in \mathbb{R}^C$ denotes the response of the channel-wise attention module. We borrow the design of channel-wise attention module from CBAM~\cite{woo2018cbam}:
\begin{equation}\label{eq:3}
    \begin{aligned}
    \mathbf{m}(F) =&\;\sigma \big(W_1(W_0({\rm pool}_{sa} (F))) \\
    &+W_1(W_0({\rm pool}_{sm} (F)))\big),
    \end{aligned}
\end{equation}
where $\sigma(\cdot)$ denotes the sigmoid function, $W_0\!\in\!\mathbb{R}^{C/r \times C}$ and $W_1\!\in\!\mathbb{R}^{C/r \times C}$, $r$ is a pre-defined positive integer controlling the ratio of dimension reduction.
An activation function ReLU($\cdot$) is adopted after the first fully connected layer $W_0$, and we omit it to simplify the notation.
${\rm pool}_{sa}(\cdot)$ and ${\rm pool}_{sm}(\cdot)$ denote spatial average-pooling and spatial max-pooling respectively. 

\subsection{Cross-domain feature augmentation}
\label{subsec:fr}

As shown in \figurename~\ref{fig:framework}, for a person image $i$ from the source domain $\mathcal{D}_{s}$ and a person image $j$ from the target domain $\mathcal{D}_{t}$, we first extract their feature vectors $\mathbf{f}_{i}^{s}$ and $\mathbf{f}_{j}^{t}$. Following Eq.~(\ref{eq:1}) and Eq.~(\ref{eq:2}), we then decompose each of them into a domain-shared identity base feature $\mathbf{b}$ and a domain-specific enhancement feature $\mathbf{e}$:
\begin{equation}\label{eq:4}
    \mathbf{f}_{i}^{s}=\mathbf{b}_{i}+\mathbf{e}_{i}^{s}, \quad
    \mathbf{f}_{j}^{t}=\mathbf{b}_{j}+\mathbf{e}_{j}^{t}.
\end{equation}
$\mathbf{b}_{i}$ and $\mathbf{b}_{j}$ are expected to encode domain-shared identity information as the base for images $i$ and $j$, respectively. 
Meanwhile, $\mathbf{e}_{i}^{s}$ and $\mathbf{e}_{j}^{t}$ (superscript $s$ and $t$ denote source and target domain respectively) are expected to encode domain-specific information as the enhancement.

As discussed in Section~\ref{subsec:fd}, the domain-shared features of different domains are exchangeable without damaging the feature distribution for each domain. Therefore, based on the disentangled features, we recompose them by exchanging the domain-shared features across domains to have augmented instances in the feature space:
\begin{equation}\label{eq:5}
    \mathbf{r}_{i}^{t}=\mathbf{b}_{i}+\mathbf{e}_{j}^{t}, \quad
    \mathbf{r}_{j}^{s}=\mathbf{b}_{j}+\mathbf{e}_{i}^{s}.
\end{equation}
Note that for a recomposed feature $\mathbf{r}_{i}^{t}$ or $\mathbf{r}_{j}^{s}$, its identity information is expected to be inherited from the domain-shared identity base feature while the domain information is expected to be inherited from the domain-specific enhancement feature.

\subsection{Optimization of DCDFA} 
\label{subsec:optimization}

In the above two subsections, we discussed the characteristics of the desired/ideal disentanglement and the augmentation. There are two necessary conditions when ideal disentanglement is achieved. 1) The identity of a recomposed (or original) feature is determined by the identity of the domain-shared feature; 2) The domain of a recomposed (or original) feature is determined by the domain of the domain-specific feature.



We drive the feature disentanglement and augmentation towards our desired roles by encouraging the recomposed and the original features to meet the above two conditions in the training process.
Generally, for two images of two persons from different domains, as illustrated in Figure~\ref{fig:framework}, we have two original features $\mathbf{f}_{i}^{s},\mathbf{f}_{j}^{t}$ and two recomposed features $\mathbf{r}_{j}^{s},\mathbf{r}_{i}^{t}$, constituting a set $\{\mathbf{f}_{i}^{s},\mathbf{f}_{j}^{t},\mathbf{r}_{j}^{s},\mathbf{r}_{i}^{t}\}$. Whenever the features $\mathbf{f}_{i}^{s},\mathbf{f}_{j}^{t}$ are well/ideally disentangled, they should meet the above two necessary conditions. For the original features and recomposed features (\ieno, $\{\mathbf{f}_{i}^{s},\mathbf{f}_{j}^{t},\mathbf{r}_{j}^{s},\mathbf{r}_{i}^{t}\}$), the subscript denotes the identity of the person and the superscript denotes the domain identity. Therefore, for the recomposed features and original features, their domain labels and identity labels, which are inherited from domain-specific features and domain-invariant features respectively when we assume the disentanglement is ideal, could be used as supervision to drive the learning towards the ideal disentanglement. Particularly, we propose two loss constraints, \ieno, cross-domain person ReID, and domain classification.

\noindent\textbf{Cross-domain person ReID constraint.} For each element in the set $\{\mathbf{f}_{i}^{s},\mathbf{f}_{j}^{t},\mathbf{r}_{j}^{s},\mathbf{r}_{i}^{t}\}$, when we take it as an anchor, there are one positive and two negative samples within this set based on their associated person identities.
We propose a cross-domain ReID loss $\mathcal{L}_{CID}$ by taking each element in this set as the anchor in turn to pull features of the same identity and push features of different identities, which is formulated as:
\begin{equation}\label{eq:6}
      \mathcal{L}_{CID}\!=\!\frac{1}{4}\sum\limits_{m=1}^{4}\!\!\bigg(\!\log\!\big( 1\!+\! {\rm {exp}}(-s^{+}_{m}/\tau)\sum\limits_{n=1}^{N^{-}_n}{\rm {exp}}(s^{-}_{mn}/\tau)\big) \bigg),    
\end{equation}
where $m$ indexes the elements in $\{\mathbf{f}_{i}^{s},\mathbf{f}_{j}^{t},\mathbf{r}_{j}^{s},\mathbf{r}_{i}^{t}\}$. With the $m$-th element as the anchor in this set, there are one positive pair and $N^{-}_n\!=\!2$ negative pairs.
$s^{+}_{m}$ denotes the (cosine) similarity of the corresponding positive pair, and $s^{-}_{mn}$ denotes the $n$-th negative pair of that. 
$\tau$ denotes a trainable temperature value initialized with one.

\noindent\textbf{Domain classification constraint.} 
As described above, the domain labels as supervision for $\{\mathbf{f}_{i}^{s}, \mathbf{r}_{j}^{s}, \mathbf{f}_{j}^{t},\mathbf{r}_{i}^{t}\}$ would be assigned as source (\ieno, 1), source (\ieno, 1), target (\ieno, 0), target (\ieno, 0), respectively. To drive the optimization towards our expected disentanglement, we use a cross entropy based domain classification loss on the original features and recomposed features as:
\begin{equation}\label{eq:7}
\begin{aligned}
    \mathcal{L}_{Domain}=-\frac{1}{4}&\big(\log (p(\mathbf{f}_{i}^{s}))+\log (p(\mathbf{r}_{j}^{s}))\\
    &+\log (1-p(\mathbf{f}_{j}^{t}))+\log (1-p(\mathbf{r}_{i}^{t}))\big),
\end{aligned}
\end{equation}
where $p(\cdot)$ denotes the probability of being classified as source domain with a trained domain classifier. We construct the domain classifier by simply stacking two ``FC+ReLU+Dropout'' blocks, a FC layer (with 2-dimensional output), and a softmax function sequentially, in which each ``FC+ReLU+Dropout'' block reduces the channel dimension of feature with a ratio of 8 and the dropout ratio is set to 0.1 experimentally.

The $\mathcal{L}_{CID}$ and $\mathcal{L}_{Domain}$ constrain each other in the optimization to resist trivial solutions for both re-identification and domain classification. For example, the joint optimization can avoid solution that $\mathbf{e}_{i}^{s}\!=\!0$ and/or $\mathbf{e}_{j}^{t}\!=\!0$. This is because whenever they are zeros, this prevents the optimization/reduction of the domain classification loss due to the conflict: $\mathbf{b}_i$ is assigned with ``source'' label (left top $\mathbf{f}_i^s$ in \figurename~\ref{fig:framework}) whenever it is combined with $\mathbf{e}_i^s$, while $\mathbf{b}_i$ is assigned with ``target'' label (right bottom $\mathbf{r}_i^t$) whenever it is combined with $\mathbf{e}_j^t$. The jointing learning of $\mathcal{L}_{CID}$ and $\mathcal{L}_{Domain}$ drives $\mathbf{b}_{i}$ and $\mathbf{b}_{j}$ to learn domain-shared base feature, and drives $\mathbf{e}_{i}^{s}$ and $\mathbf{e}_{j}^{t}$ to learn domain-specific enhancement feature. 

In order to capture as much discriminative information as possible for the feature before decomposition (\ieno, $\mathbf{f}$) and keep consistent with the original loss designs in our baseline scheme, we add the same basic ReID loss (\egno, triplet loss and identity loss) to the feature before decomposition, which we denote it as $\mathcal{L}_{BReID}$. On top of the clustering-based baseline, we experimentally find that adding ReID loss (triplet loss and identity loss) on the learned identity base feature, \ie $\mathbf{b}_{j}$, denoted by $\mathcal{L}_{RReID}^{b}$, improves the performance, which may play a role of regularization by explicitly encouraging identity base feature to be discriminative for ReID.

During testing, for a target image $j$, the feature $\mathbf{f}_{j}^{t}= \mathbf{b}_j + \mathbf{e}_j^t$, which consists of domain-shared base information and domain-specific enhancement information, is used for person matching. Thus, the decomposition module can be discarded and we do not introduce any computational increase at inference when compared with our baseline.


\section{Experiments}

\subsection{Implementation} 

We build a strong baseline based on the clustering-based method and incorporate our proposed DCDFA on top of the strong baseline. We elaborate on them respectively.

\begin{figure}[t]
	\begin{center}
		\includegraphics[width=1.0\linewidth]{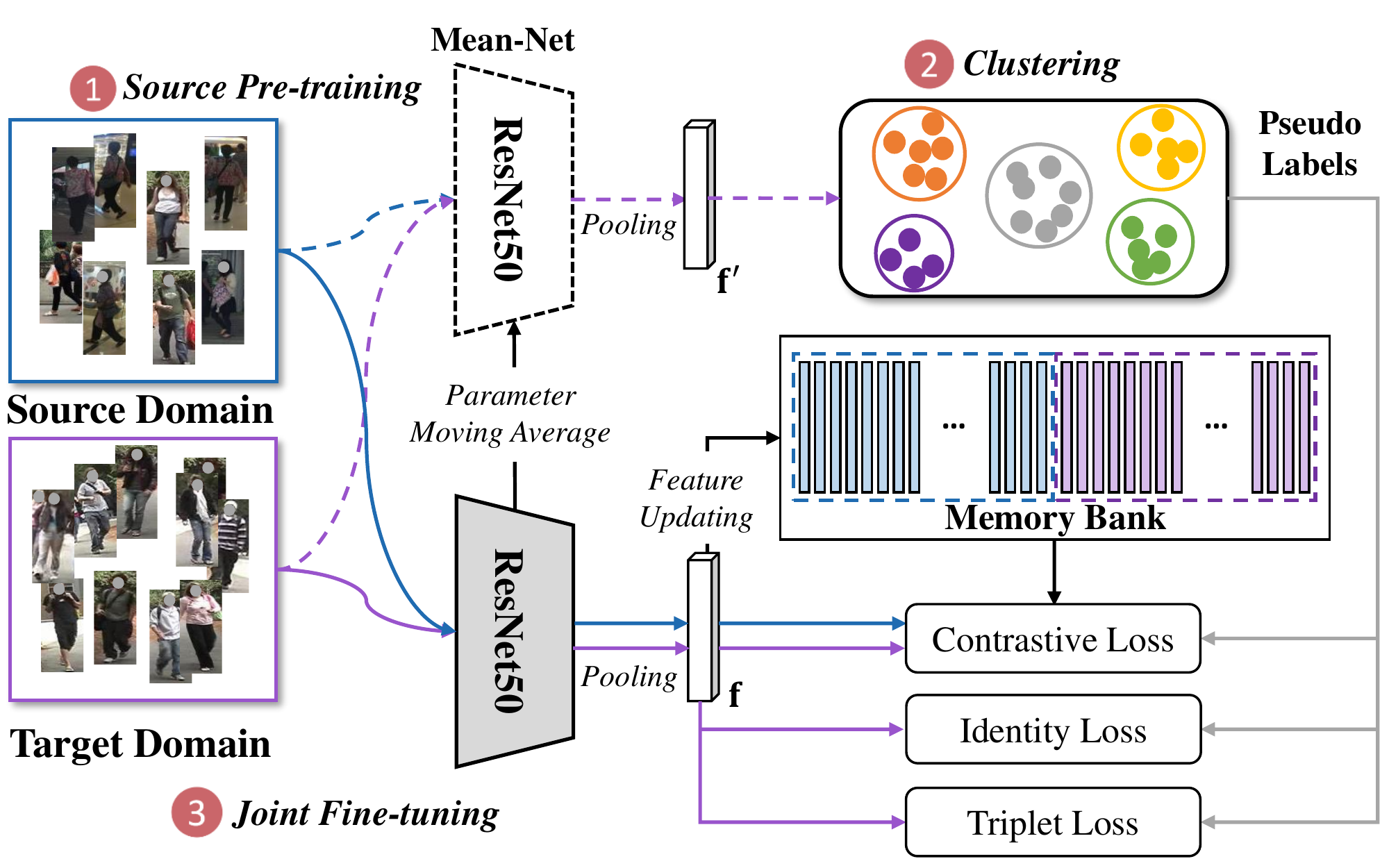}
	\end{center}
	\vspace{-2mm}
	\caption{Flowchat of our proposed strong baseline SBase for UDA person ReID. In comparison with the plain clustering-based baseline, it additionally uses the Mean-Teacher method and a memory bank (for contrastive loss). For the Mean-Teacher method, it maintains a temporal moving average (\ieno the Mean-Net) of the student network (bottom network), where the gradient back-propagation is only performed on the student network. The feature $\mathbf{f}^{'}$ from the Mean-Net is used for testing. Note that our DCDFA  will be applied on the extracted feature $\mathbf{f}$.
	}
	\label{fig:baseline}
	\vspace{-2mm}
\end{figure}

\noindent\textbf{Baseline.} We follow the general pipeline of clustering-based UDA person ReID methods \cite{fan2018unsupervised, song2020unsupervised, jin2020global,zhao2020unsupervised} to build our baseline, which consists of three main stages, \ieno, network pre-training, assignment of pseudo labels by clustering, and network fine-tuning. As illustrated in Figure~\ref{fig:baseline}, we first pre-train the network using source domain labeled data $\mathcal{D}_{s}$, where we add the basic ReID loss $\mathcal{L}_{BReID}$ (triplet loss and identity loss) on feature $\mathbf{f}^s$. We then perform clustering on the extracted features of the unlabeled target domain data $\mathcal{D}_{t}$ to generate pseudo labels for $\mathcal{D}_{t}$. Fine-tuning is performed by adding the basic ReID loss on $\mathbf{f}^t$ supervised with the predicted pseudo labels. Clustering and fine-tuning are performed iteratively.

\noindent\textbf{Strong baseline (SBase).} We incorporate the Mean-Teacher method \cite{tarvainen2017mean} which maintains a temporal moving average (Mean-Net) of the student (basic) network, and memory bank mechanism \cite{he2020momentum}, to strengthen the baseline to have a strong baseline scheme \emph{SBase} for UDA person ReID. Figure \ref{fig:baseline} illustrates this clustering-based strong baseline. In the inference, only the Mean-Net is used while the student network is discarded. We will present the ablation study of each component in Section~\ref{subsec:ablation}.



We build a memory bank and update it with the prototypes (\ieno each prototype is an averaged feature of each person) of $\mathcal{D}_{s}$, and instance features of $\mathcal{D}_{t}$. In the joint fine-tuning stage, given a query sample from $\mathcal{D}_{s}$ or $\mathcal{D}_{t}$, we add contrastive loss (with re-weighting as in circle loss \cite{sun2020circle}) w.r.t. this query and elements in the memory bank to enlarge the within-identity similarity and reduce the cross-identity similarity. Memory bank allows the collection of sufficient hard negative pairs across more mini-batches for network optimization~\cite{wang2020cross}.


\noindent\textbf{Ours (SBase+DCDFA).} We validate the effectiveness of our proposed DCDFA method on top of the strong baseline (SBase). Given a mini-batch consisting of $N$ source images and $N$ target images, for simplicity, we randomly select $N$ source-target pairs where each source image appears in exactly one pair and so does each target image. For each pair, as illustrated in \figurename~\ref{fig:framework}, we add the proposed two loss constraints for optimization. $N$ is set to 64.


\noindent\textbf{Implementation details.} We use ResNet50 pretrained on ImageNet as our backbone networks. Following the good practices in person ReID~\cite{luo2019bag}, we perform widely used data augmentation of cropping, flipping, and random erasing~\cite{zhong2020random} on the images in all experiments.
For source pre-training, each mini-batch contains 64 images of 4 identities. For our fine-tuning stage using both source and target data, each mini-batch contains $N$=64 source-domain images of 4 identities and $N$=64 target-domain images of 4 pseudo identities, in which there are 16 images for each identity. All images are resized to 256$\times$128. Similar to \cite{ge2020mutual,yang2019selfsimilarity}, we use the clustering algorithm of DBSCAN, where the maximum distance between neighbors is set to $eps=0.6$ and the minimal number of neighbors for a dense point is set to 4. 
Adam optimizer is adopted. 
The initial learning rate is set to 0.00035. The hyperparameter $r$ (see around Eq.~({\ref{eq:3}})) is set to 8. To balance the relative importance of different losses (\egno, $\mathcal{L}_{CID}$, $\mathcal{L}_{Domain}$, $\mathcal{L}_{BReID}$), following \cite{bergstra2012random}, we set the balancing weights to make the gradients of different loss items lie in a similar range and found weights of 1 work well. More details can be found in the Supplementary.

\noindent\textbf{Datasets and evaluation metrics.} We conduct experiments using three popular person ReID datasets: DukeMTMC-reID~\cite{dukemtmc} (abbreviated as ``Duke'' ), Market-1501~\cite{market} (abbreviated as ``Market'' ), and MSMT17~\cite{wei2018person}. DukeMTMC-reID~\cite{dukemtmc} has 36,411 images, where 702 identities are used for training and 702 identities for testing.
Market-1501~\cite{market} contains 12,936 images of 751 identities for training and 19,281 images of 750 identities for testing.
MSMT17~\cite{wei2018person} contains 126,441 images, where 1,041 and 3,060 identities are used for training and testing respectively. 
We use the commonly used settings, \ieno, Market $\to$ Duke, Duke $\to$ Market, Market $\to$ MSMT17, Duke $\to$ MSMT17 for evaluation.
We adopt mean average precision (mAP) and CMC Rank-1/5/10 (R1/R5/R10) accuracy (\%) for evaluation.

\subsection{Ablation study}
\label{subsec:ablation}

On top of a strong baseline (\emph{SBase}), we validate the effectiveness of our DCDFA and also study the effectiveness of each technical component in \emph{SBase}.



\vspace{-2mm}
\subsubsection{Effectiveness of components in strong baseline}
We adopt two technologies, \ieno, Mean-Teacher method (Mean-Net), memory bank (MB), to improve the plain baseline of UDA person ReID to be a strong baseline, which is denoted by ``SBase''. \tablename~\ref{tab:baseline} shows the comparisons. 1) Memory bank (MB) brings improvement of 1.7\%/1.7\% and 3.9\%/0.8\% in mAP/Rank-1 on Market$\to$Duke and Duke$\to$Market, respectively. 2) Mean-Teacher component (Mean-Net) produces 3.0\%/3.2\% and 4.0\%/0.7\% improvements in mAP/Rank-1 on Market$\to$Duke and Duke$\to$Market, respectively. 3) Our strong baseline \emph{SBase} applies both components and delivers better results.


	\begin{table}[t]
		\centering
		\footnotesize
		\begin{center}
			\begin{tabular}{|P{3.45cm}|C{0.67cm}C{0.67cm}|C{0.67cm}C{0.67cm}|}
				\hline
				\multicolumn{1}{|c|}{\multirow{2}{*}{Methods}} & \multicolumn{2}{c|}{Market$\to$Duke}&\multicolumn{2}{c|}{Duke$\to$Market}   \\
				\cline{2-5}
				\multicolumn{1}{|c|}{}& mAP & R1 & mAP & R1  \\ 
				\hline 
				Supervised learning & 72.4 & 86.0 & 83.0 & 94.2   \\
				\hline
				Baseline (Base) & 60.4 & 75.9 & 68.2 & 87.9  \\ 
				Base+MB & 62.1 & 77.6 & 72.1 & 88.7   \\ 
				Base+Mean-Net & 63.4 & 79.1 & 72.2 & 88.6 \\
				Base+MB+Mean-Net~(\textbf{SBase}) & 64.8 & 79.7 & 75.4 & 89.8 \\
				\hline
			\end{tabular}
		\end{center}
		\caption{Ablation study of components in our clustering-based strong baseline. Here, ``MB'' denotes memory bank and Mean-Net denotes the Mean-Teacher method. ``SBase'' refers to our strong baseline with both ``MB'' and Mean-Net.}
		\label{tab:baseline}
	\end{table}

	\begin{table}[t]
		\centering
		\footnotesize
		\begin{center}
			\begin{tabular}{|P{3.4cm}|C{0.7cm}C{0.7cm}|C{0.7cm}C{0.7cm}|}
				\hline
				\multicolumn{1}{|c|}{\multirow{2}{*}{Methods}} & \multicolumn{2}{c|}{Market$\to$Duke}&\multicolumn{2}{c|}{Duke$\to$Market}   \\
				\cline{2-5}
				\multicolumn{1}{|c|}{}& mAP & R1 & mAP & R1  \\ 
				\hline 
				SBase & 64.8 & 79.7 & 75.4 & 89.8 \\
				\hline
				SBase+DCDFA (using $\mathbf{b}_{j}$) & 63.9 & 78.9 & 75.1 & 89.3 \\
				SBase+DCDFA & \textbf{68.9} & \textbf{82.2} & \textbf{78.6} & \textbf{91.5} \\
				\hline
			\end{tabular}
		\end{center}
		\caption{Effectiveness of our proposed DCDFA on top of \emph{SBase}.}
		\label{tab:EffDCDFA}
	\end{table}

\vspace{-1mm}
\subsubsection{Effectiveness of our proposed DCDFA}
\label{subsubsec:effective-DCDFA}

As shown in \tablename~\ref{tab:EffDCDFA}, compared with the strong baseline \emph{SBase}, thanks to our disentanglement-based feature augmentation, our scheme \emph{SBase+DCDFA} achieves \textbf{4.1\%} and \textbf{3.2\%} improvements in terms of mAP on the two settings (Market$\to$Duke and Duke$\to$Market), which demonstrates the effectiveness of our DCDFA.

One may wonder how about the performance of our scheme when we use the domain-shared base feature alone, \ieno, $\mathbf{b}_{j}$, for inference, rather then using $\mathbf{f}_{j}^t$, which constitutes of domain-shared base feature and domain-specific enhancement feature. Table~\ref{tab:EffDCDFA} shows that \emph{SBase+DCDFA} significantly outperforms \emph{SBase+DCDFA} (using $\mathbf{b}_{j}$) by 5.0\% and 3.5\% in mAP on Market$\to$Duke and Duke$\to$Market, respectively. This indicates the domain-specific enhancement features also contain helpful discriminative information for ReID. Our joint exploration of them by generating ``ideal" augmentation is an effective strategy for improving performance of UDA.
Moreover, we found using only domain-shared feature is even poorer than the baseline \emph{SBase}. That may be because some identity discriminative information is lost when using only domain-shared features and thus this brings performance loss, even though the domain gap is alleviated in the domain-shared feature. On top of our strong baseline \emph{SBase}, we also implemented the classical adversarial domain adaption method \cite{ganin2015dann} which aims to learn domain shared/invariant features for adaptation. Similarly, the performance is close to that of \emph{SBase}, which outperforms \emph{SBase} only by 0.1\%/0.5\% in mAP on Market$\to$Duke/Duke$\to$Market.  



\begin{table}[t]
		\centering
		\footnotesize
		\begin{center}
		\resizebox{0.48\textwidth}{!}{
			\begin{tabular}{|P{4.2cm}|C{0.5cm}C{0.5cm}|C{0.5cm}C{0.5cm}|}
				\hline
				\multicolumn{1}{|c|}{\multirow{2}{*}{Methods}} & \multicolumn{2}{c|}{Market$\to$Duke}&\multicolumn{2}{c|}{Duke$\to$Market}   \\
				\cline{2-5}
				\multicolumn{1}{|c|}{}& mAP & R1 & mAP & R1  \\ 
				\hline
				SBase & 64.8 & 79.7 & 75.4 & 89.8 \bigstrut[t] \\
				SBase+$\mathcal{L}_{CID}$ & 66.1 & 80.2 & 76.0 & 89.9 \\
				SBase+$\mathcal{L}_{CID}$+$\mathcal{L}_{RReID}^{b}$ & 66.6 & 81.4 & 76.3 & 91.0 \\ 
				SBase+$\mathcal{L}_{CID}$+$\mathcal{L}_{Domain}$ & 67.5 & 81.7 & 77.2 & 91.1 \\
				SBase+$\mathcal{L}_{CID}$+$\mathcal{L}_{Domain}$+$\mathcal{L}_{RReID}^{b}$ & \textbf{68.9} & \textbf{82.2} & \textbf{78.6} & \textbf{91.5} \\
				\hline
			\end{tabular}}
		\end{center}
		\caption{Ablation study for loss function designs in our proposed DCDFA. $\mathcal{L}_{CID}$ and $\mathcal{L}_{Domain}$ denote the proposed cross-domain ReID loss and domain classification loss for original features and recomposed features across domains, respectively. $\mathcal{L}_{RReID}^{b}$ denotes the basic ReID loss (triplet loss and identity loss) added on learned identity base feature $\mathbf{b}_{j}$ for regularization.}
		\label{tab:ablation_loss}
 		\vspace{-2mm}
	\end{table}

    \begin{table*}[t!]
		\footnotesize
		\centering
		\begin{center}
			\begin{tabular}{|P{3.0cm}C{1.5cm}|C{1.0cm}C{1.0cm}C{1.0cm}C{1.0cm}|C{1.0cm}C{1.0cm}C{1.0cm}C{1.0cm}|}
				\hline
				\multicolumn{2}{|c|}{\multirow{2}{*}{Methods}} & \multicolumn{4}{c|}{Market1501$\to$DukeMTMC} & \multicolumn{4}{c|}{DukeMTMC$\to$Market1501} \\
				\cline{3-10}
				\multicolumn{2}{|c|}{} & mAP & R1 & R5 & R10 & mAP & R1 & R5 & R10 \\ 
				\hline 
				MMFA~\cite{lin2018multi} & BMVC'18 & 24.7 & 45.3 & 59.8 & 66.3 & 27.4 & 56.7 & 75.0 & 81.8 \\
				ATNet~\cite{liu2019adaptive} & CVPR'19 & 24.9 &45.1 &59.5& 64.2 & 25.6& 55.7& 73.2& 79.4\\
				SPGAN+LMP~\cite{deng2018image} & CVPR'18 & 26.2&46.4 &62.3 &68.0 &26.7 &57.7 &75.8 &82.4 \\
				CFSM~\cite{chang2018disjoint} & AAAI'19 & 27.3 & 49.8 &- & - & 28.3 & 61.2 & - & -  \\
				BUC~\cite{lin2019aBottom} & AAAI'19 & 27.5 & 47.4 & 62.6 & 68.4 & 38.3 & 66.2 & 79.6 & 84.5 \\
				ECN~\cite{zhong2019invariance} & CVPR'19 & 40.4 & 63.3 & 75.8 & 80.4 & 43.0 & 75.1 & 87.6 & 91.6 \\
				UCDA~\cite{qi2019novel} & ICCV'19 & 31.0 & 47.7 & - & - & 30.9 & 60.4 & - & - \\
				PDA-Net~\cite{li2019cross} & ICCV'19 & 45.1 & 63.2 & 77.0 & 82.5 & 47.6 & 75.2 & 86.3 & 90.2 \\
				PCB-PAST~\cite{zhang2019self} & ICCV'19 & 54.3 & 72.4 & - & - & 54.6 & 78.4 & - & - \\
				SSG~\cite{yang2019selfsimilarity} & ICCV'19 & 53.4 & 73.0 & 80.6 & 83.2 & 58.3 & 80.0 & 90.0 & 92.4  \\
				ACT~\cite{Yang2019Asymmetric} & AAAI'20 & 54.5 & 72.4 & - & - & 60.6 & 80.5 & - & - \\
				MPLP~\cite{WANG2020cvpr1} & CVPR'20 & 51.4&72.4 &82.9& 85.0 & 60.4 &84.4 &92.8& 95.0  \\
			    DAAM~\cite{Huang2020aaai} & AAAI'20 &  48.8 & 71.3 &82.4 &86.3 & 53.1 & 77.8 & 89.9 & 93.7 \\
				AD-Cluster~\cite{zhai2020adcluster} & CVPR'20 & {54.1} & {72.6} & {82.5} & {85.5} & {68.3} & {86.7} & {94.4} & {96.5} \\ 
				DIM+GLO~\cite{liu2020domain} &MM'20 &58.3 & 76.2 &85.7 & 88.5 & 65.1 & 88.3 & 94.7 & 96.3 \\
				MMT~\cite{ge2020mutual} & ICLR'20 & {65.1} & {78.0} & 88.8 & \underline{92.5} & 71.2 & {87.7} & {94.9} & {96.9} \\
				NRMT~\cite{zhao2020unsupervised} & ECCV'20 & 62.2& 77.8& 86.9& 89.5& 71.7& 87.8& 94.6& 96.5 \\				
				B-SNR+GDS-H~\cite{jin2020global} & ECCV'20 & 59.7 & 76.7 &- &- & 72.5 & 89.3 & - & - \\	
				MEB-Net~\cite{zhai2020multiple} & ECCV'20 & 66.1 & 79.6 & 88.3& 92.2 & 76.0 & 89.9 & 96.0 & 97.5\\
				SpCL~\cite{ge2020selfarXiv} & arXiv'20 & \underline{68.8} & \textbf{82.9} & \underline{90.1} & \underline{92.5} & \underline{76.7} & \underline{90.3} & \underline{96.2} & \underline{97.7} \\
				\hline
				SBase+DCDFA & Ours & \textbf{68.9} & \underline{82.2} & \textbf{91.2} & \textbf{93.0} & \textbf{78.6} & \textbf{91.5} & \textbf{96.5} & \textbf{97.7} \\
				
				\hline 
			\end{tabular}\\
			
			\begin{tabular}{|P{3.0cm}C{1.5cm}|C{1.0cm}C{1.0cm}C{1.0cm}C{1.0cm}|C{1.0cm}C{1.0cm}C{1.0cm}C{1.0cm}|}
				\hline
				\multicolumn{2}{|c|}{\multirow{2}{*}{Methods}} & \multicolumn{4}{c|}{Marke1501$\to$MSMT17} & \multicolumn{4}{c|}{DukeMTMC$\to$MSMT17} \\
				\cline{3-10}
				\multicolumn{2}{|c|}{} & mAP & R1 & R5 & R10 & mAP & R1 & R5 & R10 \\ 
				\hline 
				ECN~\cite{zhong2019invariance} & CVPR'19 & 8.5 & 25.3 & 36.3 & 42.1 & 10.2 & 30.2 & 41.5 & 46.8 \\
				SSG~\cite{yang2019selfsimilarity} & ICCV'19 & 13.2 & 31.6 &- & 49.6 & 13.3 & 32.2 & - & 51.2 \\
				DAAM~\cite{Huang2020aaai} & AAAI'20 & {20.8} & { 44.5} & {-} & {-} & { 21.6} & { 46.7} & {-} & {-} \\
				NRMT~\cite{zhao2020unsupervised} & ECCV'20 & 19.8& 43.7& 56.5 &62.2& 20.6 &45.2& 57.8& 63.3 \\
				MMT~\cite{ge2020mutual} & ICLR'20 & 22.9 & 49.2 & 63.1 & 68.8 & \underline{23.3} & \underline{50.1} & \underline{63.9} & \underline{69.8} \\
				SpCL~\cite{ge2020self} & NeurIPS'20 & \textbf{26.8} & \textbf{53.7} & \underline{65.0} & \underline{69.8} & - & - & - & - \\
				\hline	
				SBase+DCDFA & Ours & \underline{25.9} & \underline{52.8} & \textbf{65.3} & \textbf{70.1} & \textbf{27.1} & \textbf{54.3} & \textbf{67.8} & \textbf{70.7} \\
				\hline
				
			\end{tabular}
		\end{center}
		\caption{Performance (\%) comparison with the state-of-the-art methods for Unsupervised Domain Adaptation (UDA) person ReID on DukeMTMC-reID, Market-1501 and MSMT17 datasets. We mark the second-best results by \underline{underline} and the best results by \textbf{bold} text.
		}
 		\vspace{-2mm}
		\label{tab:sota}
	\end{table*}

\subsubsection{Ablation study for proposed loss designs}
We conduct ablation studies to evaluate the effectiveness of our proposed loss functions for optimizing DCDFA. \tablename~\ref{tab:ablation_loss} shows the results.
We observe that adopting $\mathcal{L}_{CID}$ only on top of SBase improves the performance slightly, while jointly using both $\mathcal{L}_{CID}$ and $\mathcal{L}_{Domain}$ significantly improves SBase by 2.7\% and 1.8\% in mAP on Market$\to$Duke and Duke$\to$Market, respectively. $\mathcal{L}_{CID}$ tends to derive domain-shared identity information into $\mathbf{b}_{j}$ but there still lacks an explicit constraint to push domain-specific information into $\mathbf{e}_{j}^{t}$. Adding $\mathcal{L}_{Domain}$ helps to promote $\mathbf{e}_{i}^{s}$ and $\mathbf{e}_{j}^{t}$ to contain domain-specific information. They jointly drive the achievement of desired feature disentanglement.

In addition, we find that adding the ReID loss (triplet loss and identity loss) \ieno, $\mathcal{L}_{RReID}^{b}$, on the learned identity base feature, \ie $\mathbf{b}_{j}$, further improves the performance. It plays a role of regularization which makes optimization easier by explicitly encouraging identity base feature to be discriminative for ReID.

\subsubsection{Ablation study for attention choices}
We use an attention module to decompose/disentangle features and find the channel-wise attention is most effective. Please see our Supplementary for more analysis and results.

\subsection{Complexity}
In testing, since our decomposition/attention module can be discarded, we do not introduce any computational increase at inference when compared with our baseline. In training, the introduction of our DCDFA only slightly increases the time complexity (\ieno, less than 5\%).

\subsection{Comparison with the state-of-the-arts}
We compare our proposed \emph{SBase+DCDFA} with the state-of-the-art methods on four domain adaptation settings in \tablename~\ref{tab:sota}.
Our \emph{SBase+DCDFA} achieves the best performance when compared with the state-of-the-art methods on Market$\to$Duke and Duke$\to$Market in mAP accuracy. On Market$\to$MSMT, our mAP accuracy is lower than SpCL~\cite{ge2020self} by 0.9\%. SpCL introduces a self-paced method which gradually creates more reliable clusters to refine the hybrid memory and learning targets. Our DCDFA, as a feature augmentation strategy, is conceptually complementary to it. We believe that applying the DCDFA on top of SpCL would further improve its performance.
SSG~\cite{yang2019selfsimilarity} performs multiple clustering on both global body and local body parts. DAAM~\cite{Huang2020aaai} learns domain-invariant features for person ReID. As discussed before, using domain-invariant feature alone is inadequate since some discriminative information is domain-specific. 
MMT~\cite{ge2020mutual} uses two networks (four models) and MEB-Net~\cite{zhai2020multiple} employs three networks (six models) to perform mutual mean teacher training, suffering high computation complexity in training. Our DCDFA uses only one network (two models) in training but still significantly outperforms MEB-Net ~\cite{zhai2020multiple}. 	



\subsection{Extension to cross-quality UDA person ReID}
We investigated the effectiveness of our proposed DCDFA on cross-quality UDA person ReID, where the person image qualities are different between the source and target domains. See our Supplementary for more details.
\section{Conclusion}
In this work, we propose a Disentanglement-based Cross-Domain Feature Augmentation (DCDFA) strategy to generate ``ideal" augmented features for training, where the augmented features characterize well the target and source domain data distributions while inheriting reliable identity labels. Particularly, we disentangle each sample feature into a domain-shared feature and a domain-specific feature, and perform cross-domain feature recomposition to increase the diversity of samples used in training.
Thanks to our designs, the recomposed features act as ``ideal" augmentation, which enables reliable inheritance of identity and approximiates the real distributions.
Experiments demonstrate the effectiveness of our DCDFA. As a feature augmentation strategy, DCDFA is generic and could be used on top of the existing UDA methods to further enhance their performance.

{\small
\bibliographystyle{ieee_fullname}
\bibliography{egbib}
}

\end{document}